# The Silver Lining Around Fearful Living

Liane Gabora

Department of Psychology, University of British Columbia
Okanagan Campus, Arts Building, 333 University Way, Kelowna BC, V1V 1V7, CANADA

**Abstract**

This paper discusses in layperson's terms human and computational studies of the impact of threat and fear on exploration and creativity. A first study showed that both killifish from a lake with predators and from a lake without predators explore a new environment to the same degree and plotting number of new spaces covered over time generates a hump-shaped curve. However, for the fish from the lake with predators the curve is shifted to the right; they take longer. This pattern was replicated by a computer model of exploratory behavior varying only one parameter, the fear parameter. A second study showed that stories inspired by threatening photographs were rated as more creative than stories inspired by non-threatening photographs. Various explanations for the findings are discussed.

We are so intimately connected to the world we live in. Not just with what is immediately around us, but also with what is happening on the other side of the globe. We can instantly communicate with people of all cultures, and even see their homes and villages on Google maps. It's fascinating. But if you think back to when you knew so much less about the world it seemed so tantalizingly full of wonder and possibility. It seems there is a tradeoff between that sense of mystery and how much you know. This got me thinking: one good thing about fear, other than it stops you from doing stupid, dangerous stuff, is that can hold you back from learning the facts inside out, which may help keep that creatively inspiring sense of wonder alive.

A STUDY OF THE IMPACT OF THREAT ON EXPLORATORY BEHAVIOR

These ideas date back to a study I carried out as a graduate student with killifish from two lakes: one with killifish predators and one without killifish predators. The fish were caught and put into a relatively small section of a large tank until they had acclimatized. This small section was separated from the rest of the tank by a door that was initially shut. The larger section contained objects that were unfamiliar to them.

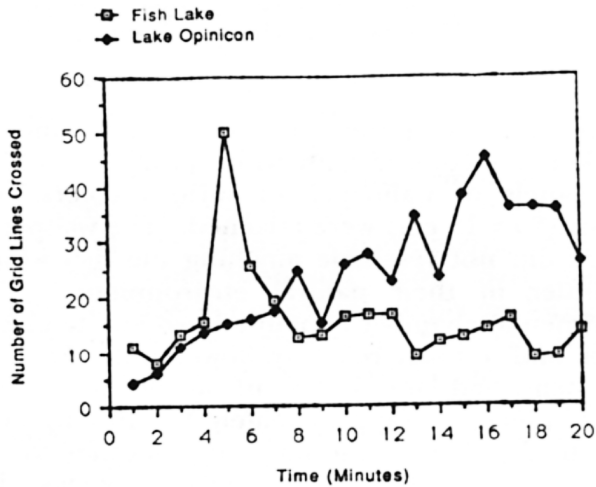

Figure 1. Total number of grid lines crossed by five fish from by Fish Lake (no predators) and Lake Opinicon (predators) per minute of exposure to a new environment.

When the door was opened and the fish were allowed to leave their "comfort zone" and enter the unfamiliar portion of the tank, both groups explored it equally thoroughly, first increasing and then decreasing the amount of exploration, and in the end covering the same amount of territory. When I plotted exploration over time I got a hump-shaped curve, and the total area under the curve was the same for both groups, as illustrated in Figure 1. However, there was a big difference between them: the fish from the lake with predators took a lot longer to explore their new surroundings than the fish from the lake without predators.

I wrote a little computer program called "Explorer" in which there were two competing desires: the desire to explore a new environment and the desire to avoid fear (Gabora & Colgan, 1990). Unless the fear was at the maximum level, Explorer would always explore a new environment until it had covered it entirely. It's exploration level would first increase as it "conquered" its fear and then decrease as there was less new territory to cover, just like what I observed with the killifish, as illustrated in Figure 2. But the higher the fear level was, the more slowly the curve rose and fell. Just by changing the fear parameter I got graphs that mirrored those of the two populations of killifish.

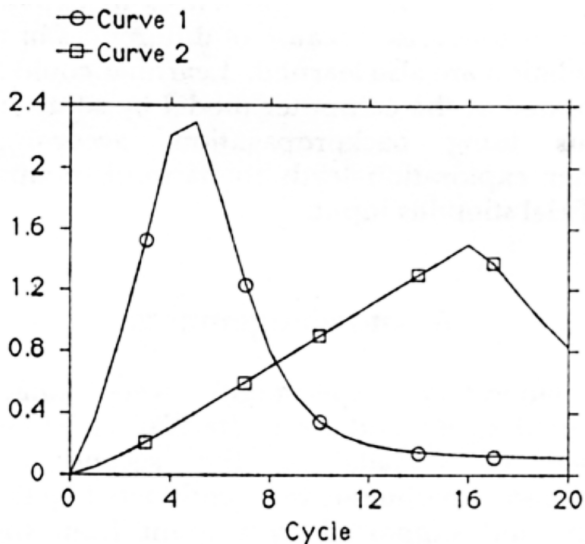

Figure 2. Exploration curves generated by two runs of the Explorer computer model. The difference between curve 1 (the Fish Lake-like curve) and Curve 2 (the Lake Opinicon-like curve) was due to varying one parameter which represented amount of fear.

I spent many long hours watching those fish, and sometimes I felt sad for the ones from the lake with predators, thinking about how their past experiences held them back. But meanwhile I myself was slowly getting to find the town I was living in a little to "known" and predictable. I remembered how when I had first arrived in that town, the roads and alleys I now knew so well had had a mystique about them. It had seemed they could lead anywhere, and that feeling of possibility was creatively inspiring. That's how I started thinking there is a tradeoff, loosely put, between knowledge and potentiality. It's a theme that, in various ways, I've been exploring ever since.

A STUDY OF THE IMPACT OF THREAT ON CREATIVE WRITING

A student of mine, Sean Riley, thought it would be interesting to test the hypothesis that situations that are demanding, threatening, or involve conflict, put one in a more creative state of mind. There was indirect evidence in the literature that this is the case. For example, it had been shown that individuals who are in the midst of conflict set broader and more inclusive cognitive categories (De Dru, Carsten & Nijstad, 2008). Creativity is also positively correlated with group conflict (Troyer & Youngreen, 2009) anxiety (Carlsson, 2002). There was also some indication that negative affect leads to greater creative output (Akinola & Mendes, 2008).

We tested the hypothesis that threatening situations put one in a more creative state of mind by conducting a study with 60 participants, students at the University of British Columbia (Riley & Gabora, 2012). First the participants viewed a series of photographs and rated each one with respect to how threatening they found it. Next they then wrote two short stories: one based on the photograph they rated as most threatening, and the other based on the photograph they rated as least threatening.

The creativity level of the stories was assessed by multiple judges, all published authors of works of fiction, who were naïve as to the purpose of the study and who were not shown or told about any of the photographs. What we found was that the stories based on threatening pictures were rated significantly higher in creativity than the stories based on non-threatening pictures.

DISCUSSION

So here in my own lab we had evidence that even just looking at threatening pictures enhances creative output! Why would this be?

My own inclination is to explain it in terms of potentiality. The non-threatening photographs depicting situations that would likely continue to unfold along predictable, mundane paths. In contrast, the threatening photographs depicting situations where there are a wide variety of possible outcomes, ranging from death to 'happily ever after', and a greater variety of feelings that would be generated by these different outcomes. Another way of saying this is that the potentiality of threatening situations is higher -- there is more at stake -- and having something at stake is vital to good story telling.

Another way to explain this is in terms of the role of the creative process in reconciling potentially stressful inconsistencies in our worldview (Curl, 2008; Gabora, 1999). We want to believe that the world is just and fair, and that we, and those we empathize with, are deserving of, and will in the end receive, just and fair treatment. A threatening stimulus confronts and contradicts this view of how the world operates. In response, we tap into our creative potential and hone in on an explanation for the threat's existence. This is done in an attempt to reconcile the worldview dichotomy, and impose a sense of meaning and understanding as to why this negative reality exists, ultimately forging a new and cohesive worldview structure.

This is related to Freud's belief that when we are thwarted and in a negative emotional state, we need to find solutions, so we are more inclined, form strong associations and make deep connections. Because of the way our mind encodes information as distributed patterns of activation, this enables more overlapping of concepts previously thought to be unrelated, and is conducive to creativity (Gabora, 2000). Thus there is a positive correlation between negative affect and creativity (Akinola & Mendes, 2008).

These explanations may not be mutually exclusive, and there could be a grain of truth to them all. Whatever explanation you choose, it seems to be the case that there may be a silver lining to threatening situations. In processing them you work something out or come to some kind of acceptance, and if you do this through music or words or the tools of an artist you may be left with a gift, a creative "precipitate" of this transformative process.

ACKNOWLEDGMENTS

The author is grateful for funding from the Natural Sciences and Engineering Research Council of Canada.